%% file: main.tex
\newif\ifanonymous \anonymoustrue
\anonymousfalse

\documentclass[runningheads]{llncs}
\usepackage[numbers, sort, comma, square]{natbib}
\usepackage{amsmath,amssymb,amsfonts}
\usepackage{algorithmic}
\usepackage{graphicx}
\usepackage{textcomp}
\usepackage{xcolor}
\usepackage{booktabs}       

\usepackage[caption=false,font=normalsize,labelfont=sf,textfont=sf]{subfig}

\newcommand{\colorR}[1]{\textcolor{red}{#1}}
\newcommand{\pagelimitmarker}[1]{~\\ {\colorR{\ifthenelse{\thepage>#1}{\Huge Exceeding the page limit}{\huge Within the page limit}}}~\\ {\huge{\colorR{~~Page Limit\,\,\,\,\, = #1}}}~\\ {\huge{\colorR{~~Current Page = $\thepage$}}}}

\usepackage{bcprules}
\usepackage{xr}
\usepackage{pdfpages}
\usepackage{hyperref}
\usepackage{cleveref}

\usepackage{algorithm}
\usepackage{algorithmic}

\usepackage{paralist}

\newif\ifdraftComments
\draftCommentsfalse

\def\mkDraftFn#1#2{%
  \expandafter\def\csname #1\endcsname##1{\ifdraftComments\textcolor{#2}{[#1: ##1]}\marginpar[$\longrightarrow$]{$\longleftarrow$}\fi}%
}
\mkDraftFn{KS}{red}
\mkDraftFn{TS}{blue}
\definecolor{darkgreen}{rgb}{0.0, 0.6, 0.39}
\mkDraftFn{TT}{darkgreen}
\mkDraftFn{HU}{purple}

\usepackage[normalem]{ulem}

\input{defs}

\usepackage{tikz}
\definecolor{lime}{HTML}{A6CE39}
\DeclareRobustCommand{\orcidicon}{
	\begin{tikzpicture}
	\draw[lime, fill=lime] (0,0)
	circle [radius=0.16]
	node[white] {{\fontfamily{qag}\selectfont \tiny ID}};
	\draw[white, fill=white] (-0.0625,0.095)
	circle [radius=0.007];
	\end{tikzpicture}
	\hspace{-2mm}
}
\def\orcidID#1{\href{http://orcid.org/#1}{\protect\raisebox{-1.25pt}{\protect\orcidicon}}}

\begin{document}

\title{Learning Heuristics for Template-based CEGIS of Loop Invariants with Reinforcement Learning}
\titlerunning{Learning Heuristics for Template-based CEGIS with Reinforcement Learning}




\ifanonymous

\author{omitted for double-blind reviewing}

\else

\author{Minchao Wu\inst{1}
\and
Takeshi Tsukada\inst{2}\orcidID{0000-0002-2824-8708} \and
Hiroshi Unno\inst{3,4}\orcidID{0000-0002-4225-8195} \and
Taro Sekiyama\inst{5}\orcidID{0000-0001-9286-230X} \and
Kohei Suenaga\inst{6}\orcidID{0000-0002-7466-8789}
}
\authorrunning{M. Wu et al.}


\institute{
  Australian National University \& Data61, CSIRO, Australia
  \email{Minchao.Wu@anu.edu.au}
  \and
  Chiba University, Japan
  \email{tsukada@math.s.chiba-u.ac.jp}\\
  \and
  University of Tsukuba, Japan
  \email{uhiro@cs.tsukuba.ac.jp}\\
  \and
  RIKEN AIP, Japan \\
  \and
  National Institute of Informatics, Japan
  \email{tsekiyama@acm.org}\\
  \and
  Kyoto University
  \email{ksuenaga@fos.kuis.kyoto-u.ac.jp}
}

\fi 

\maketitle

\input{abstract}

\input{introduction}
\input{preliminaries}
\input{CEGIS}
\input{MDP}

\input{experiments}

\section{Related Work}
\label{sec:relatedwork}
\input{related}
\input{conclusion}

\ifanonymous\else
\section*{Acknowledgments}
 This work was supported in part by
 JSPS KAKENHI Grant Numbers JP19K22842, JP19H04084, JP20H04162, and JP20H05703,
 JST CREST Grant Number JPMJCR2012, Japan, and
 ERATO HASUO Metamathematics for Systems Design Project (No.\ JPMJER1603), JST.
\fi


\clearpage
\bibliographystyle{splncs04}
\bibliography{abbrv,main,prog_lang}















\end{document}

%% file: defs.tex
\newcommand\set[1]{\{#1\}}

\renewcommand\vec[1]{\mathbf{#1}}
\renewcommand{\algorithmicrequire}{\textbf{Input:}}
\renewcommand{\algorithmicensure}{\textbf{Output:}}

\DeclareMathOperator{\CWHILE}{\mathbf{while}}
\DeclareMathOperator{\CDO}{\mathbf{do}}
\DeclareMathOperator{\CDONE}{\mathbf{done}}

\newcommand\PRE{\mathit{pre}}
\newcommand\POST{\mathit{post}}
\newcommand\SMT{\mathit{SMT}}
\newcommand\SAT{\mathit{Sat}}
\newcommand\UNSAT{\mathit{Unsat}}
\newcommand\CHANGETEMPL{\mathit{ChangeTempl}}
\renewcommand{\algorithmiccomment}[1]{{\color{blue}{\textit{#1}}}}
\newcommand\VALIDATOR{\mathit{Validator}}
\newcommand\VALID{\mathit{Valid}}
\newcommand\CEX{\mathit{Cex}}
\newcommand\SOL{\mathit{sol}}

\newcommand\UC{\mathit{uc}}

\def\preprocrlpcsat18{\ensuremath{\text{NewQL}^{\textsc{18}}}}

\def\bench2018{\textsc{SyGuS2018}}

\newcommand{\Sat}{\textsc{Sat}}
\newcommand{\Unsat}{\textsc{Unsat}}
\newcommand{\Timeout}{\textsc{Timeout}}

%% file: abstract.tex
\begin{abstract}
  \emph{Loop-invariant synthesis} is the basis of program
  verification. Due to the undecidability of the problem in general, a
  tool for invariant synthesis necessarily uses heuristics. Despite
  the common belief that the design of heuristics is vital for the
  performance of a synthesizer, heuristics are often engineered by their developers based on experience and intuition, sometimes in an \emph{ad-hoc} manner.
  %
  In this work, we propose an approach to systematically learning heuristics
  for template-based CounterExample-Guided Inductive Synthesis (CEGIS) with
  reinforcement learning. As a concrete example, we implement the approach on top of PCSat, which is an invariant
  synthesizer based on template-based CEGIS. Experiments show that PCSat guided by the
  heuristics learned by our framework not only outperforms existing state-of-the-art CEGIS-based
  solvers such as HoICE and the neural solver Code2Inv, but also has slight advantages over non-CEGIS-based solvers
  such as Eldarica and Spacer in linear Constrained Horn Clause (CHC) solving.
\end{abstract}

%% file: introduction.tex
\section{Introduction}
\label{sec:introduction}

%
Static formal verification is gaining more attention owing to the increasing impact of software malfunctions.
For its application to real-world software, its performance is of paramount importance.

One of the major properties of interest is partial correctness: given a program $c$ and logical formulae $\varphi_\PRE$ and $\varphi_\POST$, deciding whether $\varphi_\PRE$ and $\varphi_\POST$ are the correct precondition and postcondition of $c$, respectively.
A verification procedure needs to either prove or disprove that, if $c$ is executed from an initial state that satisfies $\varphi_\PRE$ and terminates, then the final state satisfies $\varphi_\POST$.
%

It is known that a key to solving a partial correctness problem is discovering an appropriate \emph{loop invariant}, or simply an \emph{invariant}~\cite{Winskel1993}.
%
The importance of an invariant leads to considerable research interest in the methods for solving an \emph{invariant synthesis problem (ISP)}.

Since ISP is undecidable, a common way of solving the ISP is to heuristically search for an invariant.
For example, \emph{CounterExample-Guided Inductive Synthesis} (CEGIS)~\cite{Solar-Lezama2006,Garg2014,Unno2021}
is a popular approach to the ISP; it repetitively guesses an invariant and \emph{proves} or \emph{refutes} the correctness of the guess by using an SMT solver such as Z3~\cite{Moura2008}.  
If a guess is refuted, then CEGIS makes another guess based on the counterexample.

A CEGIS procedure typically maintains a search space for candidate invariants from which a guess is made.
For example, \emph{template-based} CEGIS~\cite{Solar-Lezama2006,Garg2014,Unno2021} expresses the search space for candidate invariants using a \emph{template}, which is a predicate that contains parameters.
A guess is obtained by instantiating the parameters with concrete values so that the instantiated guess does not contradict the counterexamples obtained so far.

If all the candidates are refuted by the counterexamples, a CEGIS procedure heuristically expands the search space.
For a template-based CEGIS procedure, if it turns out that any instantiation of the parameters contradicts the obtained counterexamples, the procedure heuristically updates the template to make it more expressive.
Despite that the choice of heuristics for updating a template affects the performance of invariant synthesis, heuristics are often engineered by developers, sometimes in an ad-hoc manner --- their design is usually based on the experience and intuition of experts, and is not systematically explored.  

We propose a novel framework that learns heuristics for template-based CEGIS for ISP using reinforcement learning (RL).
RL is a kind of machine learning techniques that aims at learning a policy of the behavior of an agent that maximizes the rewards it receives from an environment.
RL has been successfully applied to the field of formal methods recently, including loop invariant synthesis~\cite{Si2018,Si2020} and theorem proving~\cite{NEURIPS2018_55acf853,DBLP:journals/corr/abs-2102-09756}.

To this end, we reformulate template-based CEGIS as an RL problem, in which an agent issues a sequence of actions following a policy. Each action tells the underlying synthesizer how to update the template of candidate invariants given the internal state of that synthesizer.
The agent receives rewards depending on the performance of the synthesizer after executing the actions taken by the agent.
RL algorithms are then implemented to learn a policy that maximizes the rewards. Such a policy is essentially a heuristic for expanding the search space that leads to a good performance of the synthesizer.

As a concrete example, we implement our framework on top of PCSat~\cite{Satake2020,Unno2021}, which is an invariant synthesizer based on template-based CEGIS.
We then conduct experiments using problems from standard benchmarks of invariant synthesis.
The experiments show that PCSat guided by the learned heuristics not only outperforms state-of-the-art CEGIS-based
solvers such as HoICE~\cite{Champion2018a,Champion2018} and the neural solver Code2Inv~\cite{Si2020}, but also has slight advantages over non-CEGIS-based solvers
such as Eldarica~\cite{Hojjat2018} and Spacer~\cite{Komuravelli2014} in linear CHC solving.


%

\paragraph{Contribution.}

The contribution of the present paper is summarized as follows.
\begin{itemize}
\item We reformulate the template-based CEGIS for ISP as an RL problem.
  Our reformulation models the behavior of an underlying synthesizer as a Markov decision process, in which a state represents the internal state of the synthesizer and an action represents a command that expands the search space for the candidate invariants.
\item We implement our framework on top of PCSat, which is an invariant synthesizer based on template-based CEGIS, and conduct experiments. We observe that PCSat using the learned heuristics outperforms existing CEGIS-based solvers. Experiments also show that PCSat using the learned heuristics even has slight advantages over non-CEGIS-based solvers in linear CHC solving.
\end{itemize}

\paragraph{Structure of the paper.}
The rest of this paper is structured as follows.
\Cref{sec:preliminary} reviews the preliminaries;
\Cref{sec:cegisForISP} presents the template-based CEGIS procedure in detail;
\Cref{sec:method} models template-based CEGIS as a Markov decision process;
\Cref{sec:experiments} describes the experimental results and ablation studies;
\Cref{sec:relatedwork} introduces related work; and \Cref{sec:conclusion} concludes.

%% file: preliminaries.tex
\section{Preliminaries}
\label{sec:preliminary}

\subsection{Notations}
\label{sec:notation}

We write $\vec{X}$ for a finite sequence $X_1,\dots,X_n$.  We use
symbols $\varphi$ and $\psi$ for first-order formulae over integers.
We often write $\varphi(\vec{x})$ to express that $\varphi$ may depend
on the variables $\vec{x}$.
For this $\varphi$ and
integers $\vec{n}$, we write $\varphi(\vec{n})$ for the
closed formula obtained by substituting $\vec{n}$ for $\vec{x}$ in
$\varphi$.

Let \( F(\vec{x}) \) be a predicate variable.
A \emph{constraint} over \( F \), denoted by symbols
$\Phi(F(\vec{x}))$ and $\mathcal{E}(F(\vec{x}))$, is a first-order formula that contains \( F \).
For example,
$\forall x. F(x) \implies F(x-1)$ is a constraint over \( F \);
this constraint holds if we set $F$ to
$x \le 0$ whereas it does not hold if we set $F$ to $x \ge 0$.
We write $\Phi(\varphi)$ for 
the formula that is obtained by substituting
$\varphi$ for $F$ in $\Phi$.
A \emph{ground constraint} is a constraint that does not contain a term variable.
For example, \( F(0,0) \wedge (F(1,2) \Rightarrow F(2,3)) \wedge \neg F(-1,0) \)
is a ground constraint.

We use symbol $\sigma$ for a mapping from variables to integers; this
mapping represents an assignment of values to variables.  We write
$\sigma(x)$ for a value of $x$ in $\sigma$ and $\sigma(\varphi)$ for
the 
formula obtained by replacing every variable in $\varphi$ with
its values in $\sigma$.  For example, if
$\sigma := \set{ x \mapsto 0, y \mapsto 1 }$, then $\sigma(y) = 1$ and
$\sigma(x < 5 \land y > 5) = (0 < 5 \land 1 > 5)$, which is false.

\subsection{Invariant Synthesis as CHC solving}
\label{sec:chcSolving}

It is well known that an ISP is equivalent
to solving a constraint $\Phi(F(\vec{x}))$ over a
predicate variable $F$, where $\Phi(F(\vec{x}))$ is of the form
\[
  \begin{array}{l}
    \Big(\forall \vec{x}. \varphi_\PRE(\vec{x}) \implies F(\vec{x}) \Big)
  \\ {}\land
  \Big(\forall \vec{x} \vec{y}. \varphi_{\mathit{trans}}(\vec{x}, \vec{y}) \land F(\vec{x}) \implies F(\vec{y})\Big)
  \\ {}\land
  \Big(\forall \vec{x}. F(\vec{x}) \implies \varphi_\POST(\vec{x})\Big).
  \end{array}
\]
For example, consider the program $c_1$:
\[
  \CWHILE y > 0 \CDO x \leftarrow x + 1; y \leftarrow y - 1 \CDONE.
\]
with precondition $x = 0 \land y = z \land z \ge 0$ and postcondition
$x = z$.  Then, the CHC $\forall x y z.\Phi_1(F(x,y,z))$ that expresses
$F(x,y,z)$ to be a loop invariant in the program $c_1$ can be defined using the following
$\Phi_1(F(x,y,z))$

\begin{eqnarray}
  \label{eqn:c1chc}
  \begin{array}[t]{ll}
    x = 0 \land y = z \land z \ge 0 \implies F(x,y,z) & \land\\
    y > 0 \land F(x,y,z) \implies F(x+1,y-1,z) & \land\\
    y \le 0 \land F(x,y,z) \implies x = z.
  \end{array}
\end{eqnarray}
The first conjunct ensures that $F(x,y,z)$ is implied by the initial
condition $x = 0 \land y = z \land z \ge 0$; the second conjunct
ensures that the predicate $F(x,y,z)$ is preserved by one iteration of
the loop if the guard condition ($y > 0$) of the loop holds; and the
third conjunct ensures that $F(x,y,z)$ implies the postcondition
$x = z$ if the loop terminates.  The loop invariant
$x + y = z \land y \ge 0$ satisfies $\forall x y z. \Phi_1(F(x,y,z))$.

A constraint of this form is called a
\emph{linear Constrained Horn Clause (linear CHC)} (or, simply
\emph{CHC} in this paper)\footnote{The invariant synthesis problem in this paper is thus defined in terms of three constrained Horn clauses for simplicity and $\Phi$ denotes the conjunction of the three.  This does not lose generality as linear CHC with multiple predicate variables and clauses can always be converted to this form by replacing the predicate variables with a single predicate variable representing their direct sum and then merging the clauses.}.
A \emph{solution} to the above constraint
is a logical formula $\varphi_\SOL(\vec{x})$
such that \( \Phi(\varphi_\SOL) \) is true.

\subsection{Reinforcement Learning}\label{sec:rl}
A \emph{Markov decision process}, which models an environment in RL problems, is specified by the following elements:
\begin{itemize}
  \item a set $\mathcal{S}$ of \emph{states},
  \item an \emph{initial state} $s_0 \in \mathcal{S}$,
  \item a subset $\mathcal{R} \subseteq \mathbb{R}$ of reals, called \emph{rewards},
  \item a finite set $\mathcal{A}$ of \emph{actions}, and
  \item a \emph{dynamics function} $p: \mathcal{S}\times\mathcal{A}\times
        \mathcal{R}\times(\mathcal{S}\cup\{\star\}) \longrightarrow [0,1]$.
\end{itemize}
We write $p(r,s' | s,a)$ for $p(s,a,r,s')$.
The dynamics function specifies a probability distribution for each $(s,a)\in\mathcal{S}\times\mathcal{A}$,
that means,
\[
  \sum_{s' \in \mathcal{S}\cup\{\star\}} \sum_{r \in \mathcal{R}} p(r,s' | s,a) = 1,
  \quad
  \mbox{for every $(s,a)\in\mathcal{S}\times\mathcal{A}$.}
\]
$\star$ is a special symbol meaning termination;
if the current state is $s$ and the action is $a$, then $ \sum_{r \in \mathcal{R}} p(r,\star | s,a) $
is the probability of termination
at the next step.

The behaviour of the agent is described by a \emph{policy},
which is a function $\pi : \mathcal{S} \times \mathcal{A} \longrightarrow [0,1]$
such that $\sum_{a \in \mathcal{A}} \pi(s,a) = 1$ for every $s \in \mathcal{S}$.
If the current state is $s$, the agent chooses $a$ as the next action with probability $\pi(s,a)$.
We write $\pi(a|s)$ for $\pi(s,a)$.

A pair of a Markov decision process and a policy probabilistically generates a sequence
\[
  s_0, a_0, r_1, s_1, a_1, r_2, s_2, \dots,
\]
where $a_{i}$ is a sample of $\pi({-}|s_i)$ and $s_{i+1},r_{i+1}$ is a
sample drawn from the distribution specified by $p({-},{-}|s_i,a_i)$.
We assume that the interaction terminates at step $n$,
i.e.~$s_n=\star$.  Then, the \emph{return} is the sum of the rewards
$ r_1 + r_2 + \dots + r_n $
(or \( r_1 + \gamma r_2 + \dots + \gamma^{n-1} r_n \) where \( \gamma \in [0,1] \) is a \emph{discount factor}).

The goal of RL is to find a policy that maximize
the \emph{expected return}. There are many learning algorithms that can achieve this.
For example, our implementation for the experiments in \Cref{sec:experiments} uses the \emph{first-visit on-policy Monte Carlo control}~\cite{first-visit-MC} and the \emph{Advantage Actor-Critic}~\cite{NIPS1999_6449f44a}.



%% file: CEGIS.tex
\section{Template-Based CEGIS for ISP}
\label{sec:cegisForISP}

\subsection{CounterExample Guided Inductive Synthesis (CEGIS) for ISP}
\label{sec:cegis}

\emph{CounterExample-Guided Inductive Synthesis (CEGIS)}~\cite{Solar-Lezama2006} solves a given CHC
\( \Phi(F(\vec{x})) \)
via the interaction between a
\emph{synthesizer (S)} and a \emph{validator (V)}.  Its high-level workflow
is described as follows.
\begin{itemize}
\item S maintains a ground constraint $\mathcal{E}(F(\vec{x}))$;
  $\mathcal{E}(F(\vec{x}))$ is called \emph{example
    instances}.
  The constraint $\mathcal{E}(F(\vec{x}))$ is a necessary condition for \( \Phi(F(\vec{x})) \),
  i.e., every solution of \( \Phi(F(\vec{x})) \) satisfies \( \mathcal{E}(F(\vec{x})) \) as well.
  The task of S is to synthesize a
  \emph{candidate solution} $\psi(\vec{x})$, which is a solution of \( \mathcal{E}(F(\vec{x})) \).
  Notice that S
  does not access the CHC $\Phi(F(\vec{x}))$.
  S then
  sends $\psi(\vec{x})$ to V.
\item V checks whether the candidate solution $\psi(\vec{x})$ sent from S is a real solution to
  \( \Phi(F(\vec{x})) \) by checking whether
  \( \varphi_\PRE(\vec{x}) \wedge \neg \psi(\vec{x}) \), \( \varphi_{\mathit{trans}}(\vec{x}, \vec{y}) \wedge \psi(\vec{x}) \wedge \neg \psi(\vec{y}) \) or \( \psi(\vec{x}) \wedge \neg \varphi_\POST(\vec{x}) \) is satisfiable.
  Most solvers
  implement this step by querying the satisfiability of
  the above formulas
  using an off-the-shelf solver such as Z3.
  If all of the formulas are unsatisfiable, then it follows that
  $\Phi(\psi(\vec{x}))$ is true and thus $\psi(\vec{x})$ is a real solution to the CHC
  $\Phi(F(\vec{x}))$.
  Otherwise,
  one of the above formulas is satisfiable, e.g., \( \varphi_{\mathit{trans}}(\vec{c}, \vec{d}) \wedge \psi(\vec{c}) \wedge \neg \psi(\vec{d}) \) is true for some vectors \( \vec{c}, \vec{d} \) of constants.
  Then, V sends $ F(\vec{c}) \Rightarrow F(\vec{d}) $
  to S as a new example instance to be satisfied.
  S updates \( \mathcal{E}(F(\vec{x})) \) to \( \mathcal{E}(F(\vec{x})) \wedge (F(\vec{c}) \Rightarrow F(\vec{d})) \) and seeks a new candidate solution again.
\end{itemize}


\subsection{Template-Based Approach to CEGIS}
\label{sec:templateBasedCEGIS}

In each step of a CEGIS-based CHC solving, a synthesizer needs to find a
candidate solution that satisfies all the constraints in the current
example instance $\mathcal{E}(F(\vec{x}))$.  One of the strategies to implement
the candidate-solution discovery
is called a \emph{template-based} approach~\cite{Garg2014,Alur2015,Unno2021}.

A template-based synthesizer works as follows.  It maintains an example
instance $\mathcal{E}(F(\vec{x}))$ and a \emph{template}
$\psi(\vec{a},\vec{x})$,
which is a predicate over \( \vec{x} \) with \emph{parameters} \( \vec{a} \),
and constructs a candidate solution by finding an appropriate assignment to \( \vec{a} \).
An appropriate assignment to \( \vec{a} \) can be computed by using an SMT solver such as Z3:
since \( \mathcal{E}(F(\vec{x})) \) is a ground constraint, i.e., a formula with no quantifier nor variable, \( C(\vec{a}) := \mathcal{E}(\psi(\vec{a},\vec{x})) \) is a quantifier-free formula with free variables \( \vec{a} \) and an SMT solver gives a satisfying assignment \( \sigma \) to \( \vec{a} \), provided that \( C(\vec{a}) \) is satisfiable.
Then \( \psi(\sigma(\vec{a}), \vec{x}) \) is a candidate solution.
If \( C(\vec{a}) \) is not satisfiable,
there is no candidate solution of the form \( \psi(\vec{c}, \vec{x}) \), where \( \vec{c} \) is a vector of constants.
Then, the synthesizer heuristically updates the template and uses it
to discover a new candidate solution. The strategies for updating the templates is what we mean by \emph{heuristics} below.


For Constraint~\ref{eqn:c1chc},
a synthesizer would be able to discover the solution
$x + y = z \land y \ge 0$ if
it designates a template
$a_1 x + b_1 y + c_1 z \ge 0  \land
a_2 x + b_2 y + c_2 z \ge 0  \land
a_3 x + b_3 y + c_3 z \ge 0$;
the above solution is obtained
once an SMT solver finds a solution
$(a_1,b_1,c_1,a_2,b_2,c_2,a_3,b_3,c_3)=
(1,1,-1,-1,-1,1,0,0,1)$.

Notice that a template $\psi(\vec{a},\vec{x})$ determines a \emph{set}
of candidate solutions, each member of which is an instantiation of
$\vec{a}$ in $\psi(\vec{a},\vec{x})$ to concrete values.  Therefore, a
template in the template-based approach determines the search space
for a candidate solution of a given constraint.  A template is said to
be more \emph{expressive} than another if the set of formulae that is
obtained by instantiating the parameters in the latter is a subset of
the former.

There is a trade-off between the expressiveness of a template and the
efficiency of a synthesizer~\cite{Padhi2019,Unno2021}.  If one uses an
expressive template, there is more chance that there is a true
solution that can be obtained by instantiating the template.
However, the constraint $C(\vec{a})$ generated by using an
expressive template tends to be complex, which incurs performance
degradation of SMT solving and therefore a synthesizer.
In deciding which template to be used, it is crucial to find a sweet
spot that addresses this trade-off.



\begin{algorithm}[t]
  \begin{algorithmic}[1]
    \caption{Template-based synthesizer}
    \label{alg:pcsat}
    \STATE{$\psi(\vec{a},\vec{x}) \leftarrow$ the initial template} \label{line:initial}
    \STATE{$\mathcal{E}(F(\vec{x})) \leftarrow \emptyset$}

    \WHILE{Timeout is not reached}

    \STATE{$C(\vec{a}) \leftarrow \mathcal{E}(\psi(\vec{a},\vec{x}))$}
    \STATE{$r_1 \leftarrow \SMT(C(\vec{a}))$}
    \IF{$r_1 = \SAT(\sigma)$}
    \STATE{}\COMMENT{Instantiate parameters}
    \STATE{$\vec{c} \leftarrow \sigma(\vec{a})$}
    \STATE{}\COMMENT{Send the candidate solution to the validator}
    \STATE{$r_2 \leftarrow \VALIDATOR(\psi(\vec{c},\vec{x}))$}
    \STATE{}\COMMENT{Check whether $\psi(\vec{c},\vec{x})$ is a real solution}

    \IF{$r_2 = \VALID$}
    \STATE{Return $\psi(\vec{c},\vec{x})$ as a real solution}

    \ELSIF{$r_2 = \CEX(\mathcal{E'}(F(\vec{x})))$}
    \STATE{}\COMMENT{Cex is found; new example instance is received.}
    \STATE{$\mathcal{E}(F(\vec{x})) \leftarrow \mathcal{E}(F(\vec{x})) \wedge \mathcal{E'}(F(\vec{x}))$}
    \ENDIF


    \ELSIF{$r = \UNSAT(I)$}
    \STATE{$\psi(\vec{a},\vec{x}) \leftarrow \CHANGETEMPL(\psi(\vec{a},\vec{x}),I)$} \label{line:changeTempl}

    \ENDIF

    \ENDWHILE


  \end{algorithmic}
\end{algorithm}

\Cref{alg:pcsat} is a typical template-based synthesizer for CHC solving.
This procedure uses the following subprocedures:
\begin{itemize}
\item $\SMT(C(\vec{a}))$: Decides whether the predicate $C(\vec{a})$
  is satisfiable or not by an SMT solver.  If it is satisfiable, then
  it returns $\SAT(\sigma)$ where $\sigma$ is a value assignment to
  $\vec{a}$ in $C(\vec{a})$ such that $\sigma(C(\vec{a}))$ is valid.
  If not, it returns $\UNSAT(I)$, where $I$ is a collection of various
  information on the behavior of the decision procedure (e.g.,
  consumed time and memory).  $I$ also includes explanations why
  $C(\vec{a})$ is unsatisfiable such as an unsat core, which is an
  unsatisfiable subconstraint of $C(\vec{a})$.  We write $I.\UC$ for
  the unsat core contained in $I$.
\item $\VALIDATOR(\varphi(\vec{x}))$: Sends the candidate solution
  $\varphi(\vec{x})$ to the validator and let it decide whether it is a
  real solution.  If $\varphi(\vec{x})$ is indeed a solution, then the
  validator returns $\VALID$.  Otherwise, the validator returns
  $\CEX(\mathcal{E}(F(\vec{x})))$, where $\mathcal{E}(F(\vec{x}))$ is the new example
  instance to be satisfied by solutions.
\item $\CHANGETEMPL(\psi(\vec{a},\vec{x}),I)$: Heuristically updates
  the template $\psi(\vec{a},\vec{x})$ to a new one based on the
  information $I$ returned by the SMT solver.
\end{itemize}


\subsection{Synthesizer Implemented in PCSat}
\label{sec:learnerInPCSat}

PCSat~\cite{Satake2020,Unno2021} is one of the tools for CHC solving based on
template-based CEGIS.  It uses a family of template
$\psi_{\vec{N},P,Q}(\vec{a},\vec{x})$, in which each template is
determined by the values $P,Q$ and $\vec{N}=(N_1,\dots,N_M)$ (of which the length is denoted by \(M\)).
Assume that $\vec{x} := (x_1,\dots,x_L)$.
Then the parameter $\vec{a}$ used in these
templates consists of the following parameters.
\begin{itemize}
\item \emph{Coefficient parameters}
  $ a^{(ij)}_k $ for each $ 1 \le i \le M $, $1 \le j \le N_i$ and $1 \le k \le L$.
\item \emph{Constant parameters}
  $c^{(ij)}$ for each $1 \le i \le M$ and $1 \le j \le N_i$.
\end{itemize}
The template family $\psi_{\vec{N},P,Q}(\vec{a},\vec{x})$ is defined as
follows using these parameters:
\[
    \bigvee\limits_{i=1}^{M} \bigwedge\limits_{j=1}^{N_i} \sum_{k=1}^{L} a^{(ij)}_k x_k \ge c^{(ij)}
    \quad\land\quad
    \bigwedge\limits_{i=1}^{M} \bigwedge\limits_{j=1}^{N_i} \sum_{k=1}^{L} |a_k^{(ij)}| \le P
    \quad\land\quad
    \bigwedge\limits_{i=1}^{M} \bigwedge\limits_{j=1}^{N_i} |c^{(ij)}| \le Q.
\]
We explain the above definition in the following.
\begin{itemize}
\item The subformula
  $\bigvee\limits_{i=1}^{M} \bigwedge\limits_{j=1}^{N_i} \sum_{k=1}^{L} a^{(ij)}_k x_k \ge c^{(ij)}$
  is a boolean combination of linear
  inequalities $\sum_{k=1}^{L} a^{(ij)}_k x_k \ge c^{(ij)}$ expressed in a
  disjunctive normal form (DNF).  The parameter $M$ (resp.\ $N_i$) is the
  number of disjuncts (resp.\ conjuncts in each disjunct) in this
  template; therefore, the larger they are, the more expressive the
  template is.
\item The subformula
  $\bigwedge\limits_{i=1}^{M} \bigwedge\limits_{j=1}^{N_i}
  \sum_{k=1}^{L} |a_k^{(ij)}| \le P$ bounds the sum of the absolute
  values of the coefficients in each linear inequality (i.e., the
  $L^1$ norm of each $(a_k^{(ij)})_{1\le k \le L}$).  The bound $P$ is a natural
  number or $\infty$; if $P = \infty$, then the coefficients may be
  any number.  The larger $P$ is, the more expressive the template is.
\item The subformula
  $\bigwedge\limits_{i=1}^{M} \bigwedge\limits_{j=1}^{N_i} |c^{(ij)}|
  \le Q$ bounds the absolute value of the constant $c^{(ij)}$ in
  each linear inequality.  The bound $Q$ is a natural number or
  $\infty$.  The larger $Q$ is, the more expressive the template is.
\end{itemize}
The strategy of PCSat actually uses a subset of above-defined templates, namely the subset of templates of the form \( ([N]*M, P, Q) \) where \( [N]*M \) means the list of length \( M \) consisting only of \( N \).
In other words, \( N_i = N_{i'} \) holds for every template reachable by the hand-crafted strategy of PCSat.


%

As mentioned in \Cref{sec:templateBasedCEGIS}, an update to a template happens when the constraint $C(\vec{a})$ on parameters $\vec{a}$ is unsatisfiable.
PCSat decides how to update a template using the unsat core of $C(\vec{a})$.

Concretely, the heuristic implemented by PCSat is as follows.
$(M,N,P,Q)$ are initialized to $(1,1,1,0)$.
%
%
When PCSat needs to update the current template, it increments $M$ or $N$ by $1$ and it may increment $P$ and/or $Q$ depending on the unsat core.
$M$ and $N$ are incremented in alternation.
%
%
If $P$ occurs in the unsat core, then $P$ is incremented by $1$.
If $Q$ occurs in the unsat core and $Q < 3$, then $Q$ is incremented by $1$; if $Q \ge 3$, then $Q$ is set to $\infty$.

%% file: MDP.tex
\section{Finding Heuristics with Reinforcement Learning}
\label{sec:method}

This section describes how we formulate the problem of learning heuristics as a reinforcement learning (RL) problem.
%
%
In our setting, the environment is an implementation of the template-based CEGIS parameterized by heuristics.
%
We learn a policy that represents a heuristic, which tells the environment the shape of the template that should be tried in order to guess the next candidate solution.
The long-term goal is to find a solution of a given CHC, preferably in a short time.

%
We model the template-based CEGIS procedure as a Markov decision process (MDP) as follows, and implement RL algorithms to solve such an MDP.

\paragraph{States.}
A state of the MDP in our formalism is \( (\vec{N}, P, Q, f_1, f_2, z) \) where $(\vec{N},P,Q)$ are values that determine the current template, 
$f_1, f_2$ are boolean values that summarize the information of the unsat core for the current template,
and $z$ is the number of candidate solutions that the learner found since the previous update of template.
$f_1$ is true if the unsat core contains $P$, and $f_2$ is true if the unsat core contains $Q$.
The information expressed by flags \(f_1\) and \(f_2\) is also used by heuristics engineered by the PCSat developers.
The parameter \( z \) is inspired by
Code2Inv~\cite{Si2020}, which uses the number of example instances
satisfied by the current state of the environment as part of a reward.

\paragraph{Actions.}
An action is a tuple \( (\vec{n}, p, q) \in \mathbb{N}^* \times (\mathbb{N} \cup \{\infty\}) \times (\mathbb{N} \cup \{\infty\} \)), which updates the current template \( (\vec{N}, P, Q) \) to \( (\vec{N} + \vec{n}, P+p, Q+q) \)\footnote{The actions adopted in the present paper only increase the complexity of templates.  Before settling on the current design, we tried actions that can reduce the complexity of templates and states that incorporate the timeout feedback from Z3 (as a hint for the complexity of the current template).  We however omitted them in the present paper because early experiments showed that learning with Monte Carlo control became unstable and the performance decreased in such a setting.}.
Here \( \vec{N} + \vec{n} \) is the coordinate-wise sum;
if the lengths of \( \vec{N} \) and \( \vec{n} \) are different, we append \( 0 \)'s to the tail of the shorter one.
For example, if \( \vec{N} = (1,1) \) and \( \vec{n} = (0,0,1)\), then \( \vec{N} + \vec{n} = (1,1,1) \).
Note that the length \( M \) of \( \vec{N} \) can be increased by choosing sufficiently long \( \vec{n} \).

\paragraph{Rewards.}
Since simply checking whether a solution is found or not may have the problem of sparse rewards~\cite[Section~17.4]{Sutton2018} that makes learning harder, we
define the reward associated with each action to be $-T$, where $T$ is the time spent since the last invocation of the agent.
The sum of the rewards is then naturally $-T_{\mathit{total}}$, where
$-T_{\mathit{total}}$ is the total time spent for the run.
Intuitively, the earlier a solution is found, the more reward is given.
The smallest reward is given if the synthesizer eventually times out when guided by the agent.

%% file: experiments.tex
\section{Experiments}
\label{sec:experiments}

%
%
%
%
%
\paragraph{Datasets and tasks.}
We use the problems in the Inv-Track of the
SyGuS-Comp~\cite{syguscomp} 2019 competition\footnote{The problems can be found at
\url{https://github.com/SyGuS-Org/benchmarks}.} as the data set of our experiments.
A tool for this competition is supposed to return one of the
three answers: {\Sat}, indicating that the tool successfully
synthesized an answer to the given problem; \HU{precisely speaking, SyGuS-Comp does not require tools to return \Unsat}{\Unsat}, indicating
that there is no solution to the given problem; and {\Timeout}, indicating that the tool fails to solve the given problem within a specified time limit.
An answer of {\Sat} must be accompanied by a witness, which is an invariant in the case of
Inv-Track.
The Inv-Track consists of 858 problems for evaluating the performance of invariant-synthesis tools. We randomly split the problems in the Inv-Track into training and test sets in an 80:20 ratio.
%
The learning task is to train an agent that is capable of guiding PCSat to find a solution to each problem as soon as possible.

\paragraph{Configurations.}
%
We train the agent using two different reinforcement learning algorithms: the first-visit on-policy Monte Carlo (MC) control~\cite{first-visit-MC} and Advantage Actor-Critic (A2C)~\cite{NIPS1999_6449f44a}.
%
The former is a tabular method that relies on state-action value tables and the latter leverages deep learning.
The state and action spaces are as described in Section~\ref{sec:method}.

For MC, in order to control the size of the state-action value table and make the learning tractable, we set up an upper bound for each parameter in the state representation.\footnote{%
  The action space here is restricted to \( (\vec{n}, p, q) \in \{0,1\}^{\le 4} \times \{0,1,\infty\} \times \{0,1,\infty\} \).  In particular, the number of conjuncts in templates is unbounded but the number of disjuncts is at most \( 4 \).  The agent's ability to observe a state is thus limited.  For \( N_i \), the agent does not distinguish \( 4 \) with greater values; for \( P \) and \( Q \) it does not distinguish between \( 5 \) with greater values.  For example, the state \( ((2,1,4,0),7,4) \) looks \( ((2,1,\,{\ge} 4, 0), \,{\ge} 5, 4) \) to the agent, where \( {\ge}4 \) and \( {\ge}5 \) are symbols representing numbers greater than or equal to \( 4 \) and \( 5 \), respectively.}
\TS{I think it would be better to explain which numbers are set as upper bounds
for each parameter because some reviewers are concerned with reproducibility.
If we have no sufficient space, it could be shown in a separated appendix (note
that TACAS doesn't allow attaching appendices beyond the page limit to the main
paper).}
We train the agent with $\epsilon$-greedy%
  \footnote{Technical terms of reinforcement learning that are not
  explained in this paper are used
  in this and the next paragraphs, in order to
  appropriately describe the experimental setting.
  See a textbook~\cite{Sutton2018} for a reference.}
action selection with $\epsilon=0.05$ and evaluate it with the greedy policy. The discount factor is set to 1. The time limit for each problem during training is 120 seconds.

For A2C, given a trajectory $\tau=(s_0, a_0, r_0, s_1, a_1, r_1, \dots, s_T, a_T, r_T)$, our objective is to maximize the following expected policy rewards with discount factor $\gamma=0.99$
\begin{align*}
	\mathbb E_{\tau\sim\pi_{\theta}(a_t|s_{t})}\Big[\sum_{t'=t}^T \gamma^{t'-t} r_{t'}-V_{\varphi}(s_{t})\Big]
\end{align*}
where $\theta$ is the parameters of the actor policy $\pi_{\theta}$ and $\varphi$ is the parameters of the critic policy $V_{\varphi}$. Each policy is parameterized by a multi-layer perceptron with two hidden layers of dimensions 256 and 512. We use the loss $$\mathcal{L}(\varphi)=\sum\limits_{t}^T(\sum_{t'=t}^T \gamma^{t'-t} r_{t'}-V_{\varphi}(s_{t}))^2$$ to train the critic. Both policies are trained using RMSProp~\citep{tieleman2012lecture} with a learning rate of $5\times 10^{-5}$. The time limit for each problem during training is 10 seconds.
%
%
%

%
For each experiment, we train the agent for 200 epochs on the training set. Among the learned policies, we chose the one that has the best performance on the training set and evaluate them on the test set.
All the experiments are conducted on PCSat with SMT solver Z3 (version
4.8.9) on a 2.8GHz Intel(R) Xeon(R) CPU with 64 GB RAM and a Tesla V100 GPU with 16GM RAM.

%
%
%


\subsection{Evaluation}
\label{sec:evaluation}
In this section, we study the quality of the heuristics learned by our framework by comparing them with the heuristic engineered by human experts (i.e., the PCSat developers), existing non-learning-based solvers (CVC4~\cite{Barrett2011,Reynolds2019,Barbosa2019}, LoopInvGen~\cite{Padhi2019}, HoICE~\cite{Champion2018}, Eldarica~\cite{Hojjat2018}, FreqHorn~\cite{8102247} and Spacer~\cite{Komuravelli2014}) and the state-of-the-art neural solver Code2Inv~\cite{Si2018,Si2020}\footnote{We have
  also tried to use CLN2INV~\cite{DBLP:conf/iclr/RyanWYGJ20}, which is
  a deep-learning-based invariant-synthesis tool, as a baseline.
  However, we excluded it from our baseline since the implementation
  that is made public is not fully automated.  It uses
  hints that are given by human to solve certain problems.}

If classified by the underlying approaches to ISP,
LoopInvGen, HoICE, FreqHorn and Code2Inv are all CEGIS-based solvers.
LoopInvGen is based on CEGIS with greedy set covering for synthesis.
HoICE uses decision-tree based CEGIS while FreqHorn and Code2Inv use grammar-based CEGIS.
Eldarica and Spacer are non-CEGIS-based solvers. Eldarica uses CounterExample-Guided Abstraction Refinement (CEGAR) with predicate abstraction, and
Spacer is based on Property Directed Reachability (PDR).

LoopInvGen and CVC4 are respectively the winners of SyGuS-Comp 2018 and 2019.
Eldarica and Spacer are participants of some of the linear CHC tracks and won the 1st and 2nd places\footnote{According to \url{https://chc-comp.github.io/2021/presentation.pdf}. Also note that our RL formulation itself is general so that it is also applicable to non-linear CHC, and so is PCSat. The present paper focuses on the effectiveness for linear CHC as a first step because it is a non-trivial class of practical importance.}.


%



\begin{table}
  \caption{Performance of the learned heuristics and that of the baselines on the test set. \texttt{PCsat/random} refers to PCSat guided by a random policy, and \texttt{PCSat/expert} refers to PCSat using the heuristic engineered by its developers. \texttt{PCSat/A2C} and \texttt{PCSat/MC} refer to PCSat using the heuristics learned with the corresponding approaches, 
  as described in~\Cref{sec:experiments}. \texttt{PCSat/A2C+PCSat/MC} means that the two policies work jointly in parallel, and a problem is solved if one of them returns a solution.
  See footnote~\ref{footnote:reproduction} for the solvers marked with *.
\\~}
  \label{tab:baselines}
  \centering
\begin{tabular}{lccccc}
\toprule
 Methods &   approach & ~sat~ & ~unsat~ & ~timeout~ & ~time(s)~ \\
 \midrule
\texttt{FreqHorn} & CEGIS & 70 & 0 & 101 & 6863 \\
\texttt{LoopInvGen*} & CEGIS & 87 & 5 & 79 & 5086 \\
\texttt{CVC4*} & - & 102 & 9 & 60 & 3873 \\
\texttt{PCSat/random} & CEGIS & 116 & 9 & 46 & 3383 \\
\texttt{Eldarica} & CEGAR & 122 & 9 & 40 & 3714 \\
\texttt{PCSat/expert} & CEGIS & 135 & 9 & 27 & 2130 \\
\texttt{HoICE} & CEGIS & 141 & 8 & 22 & 1707 \\
\midrule
\texttt{PCSat/A2C} & CEGIS & 145 & 9 & 17 & 1947\\
\texttt{PCSat/MC} & CEGIS & 146 & 9 & 16 & 1550\\
\texttt{PCSat/A2C+PCSat/MC} & CEGIS & 149 & 9 & 13 & 1460\\
\midrule
\texttt{Spacer} & PDR & 156 & 9 & 6 & 380\\
\bottomrule
\end{tabular}
\end{table}

\begin{table}
  \caption{Performance with a time limit of 600 seconds.
\\~}
  \label{tab:baselines-600s}
  \centering
\begin{tabular}{lccccc}
\toprule
 Methods &   approach & ~sat~ & ~unsat~ & ~timeout~ & ~time(s)~ \\
 \midrule
\texttt{FreqHorn} & CEGIS & 91 & 0 & 80 & 52933 \\
\texttt{LoopInvGen*} & CEGIS & 90 & 5 & 76 & 46445 \\
\texttt{CVC4*} & - & 106 & 9 & 56 & 34869 \\
\texttt{PCSat/random} & CEGIS & 121 & 9 & 41 & 25858 \\
\texttt{PCSat/expert} & CEGIS & 145 & 9 & 17 & 12537 \\
\texttt{HoICE} & CEGIS & 147 & 9 & 15 & 10253 \\
\midrule
\texttt{PCSat/MC} & CEGIS & 150 & 9 & 12 & 8705\\
\texttt{Eldarica} & CEGAR & 151 & 9 & 11 & 12866 \\
\texttt{PCSat/A2C} & CEGIS & 151 & 9 & 11 & 8319\\
\texttt{PCSat/A2C+PCSat/MC} & CEGIS & 154 & 9 & 8 & 6413\\
\midrule
\texttt{Spacer} & PDR & 156 & 9 & 6 & 3619\\
\bottomrule
\end{tabular}
\end{table}

\paragraph{Comparison with non-learning-based solvers.}
\autoref{tab:baselines} shows the number of problems\footnote{\label{footnote:reproduction}Unfortunately, we were not able to reproduce the same level of performance as described in the competition report of LoopInvGen and CVC4, possibly due to the differences in versions and configurations used by the solvers.
CVC4 (resp.~LoopInvGen) solved 592 (resp.~512) instances out of 858 according to the competition report but only 505 (resp.~451) in our experiment.
We are not sure how many instances in the test set was solved by CVC4 and LoopInvGen in the competition, since the report just tells the total number of solved instances and does not report which instances are solved by each solver.}  in the test set solved by each method given a time limit of 60 seconds. It can be seen that the learning is effective --- compared to PCSat guided by a random policy, the policy learned by MC solves significantly more problems (146 as opposed to 116). The learned heuristics are also better than the heuristic engineered by human experts. PCSat guided by MC solves 146, while it solves only 135 when using heuristic designed by its developers.

\begin{figure*}[t] \centering
 \subfloat[\texttt{A2C} vs. \texttt{MC}]{
   \includegraphics[width=0.32\textwidth]{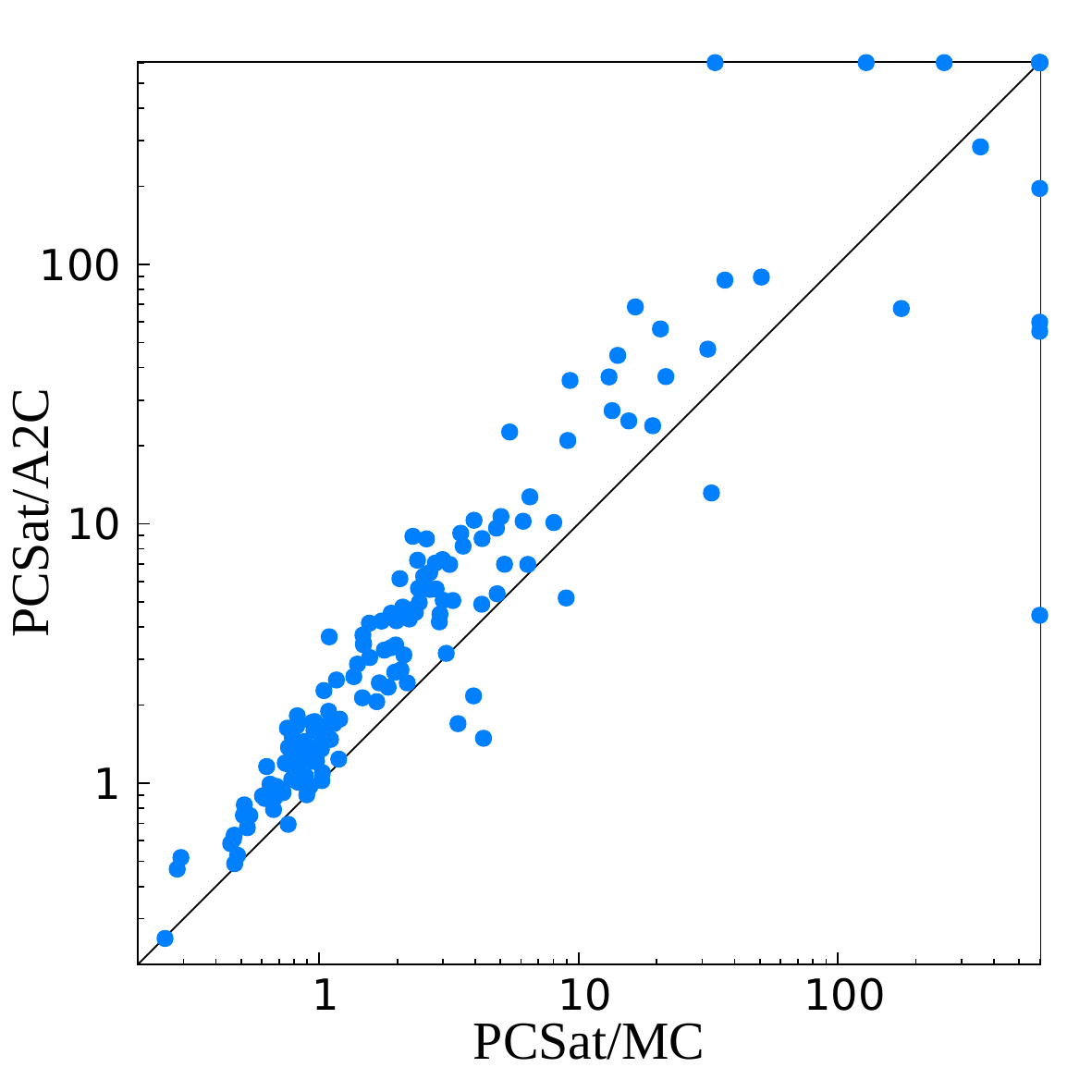}
 }
 \subfloat[\texttt{MC} vs. \texttt{Eldarica}]{
   \includegraphics[width=0.32\textwidth]{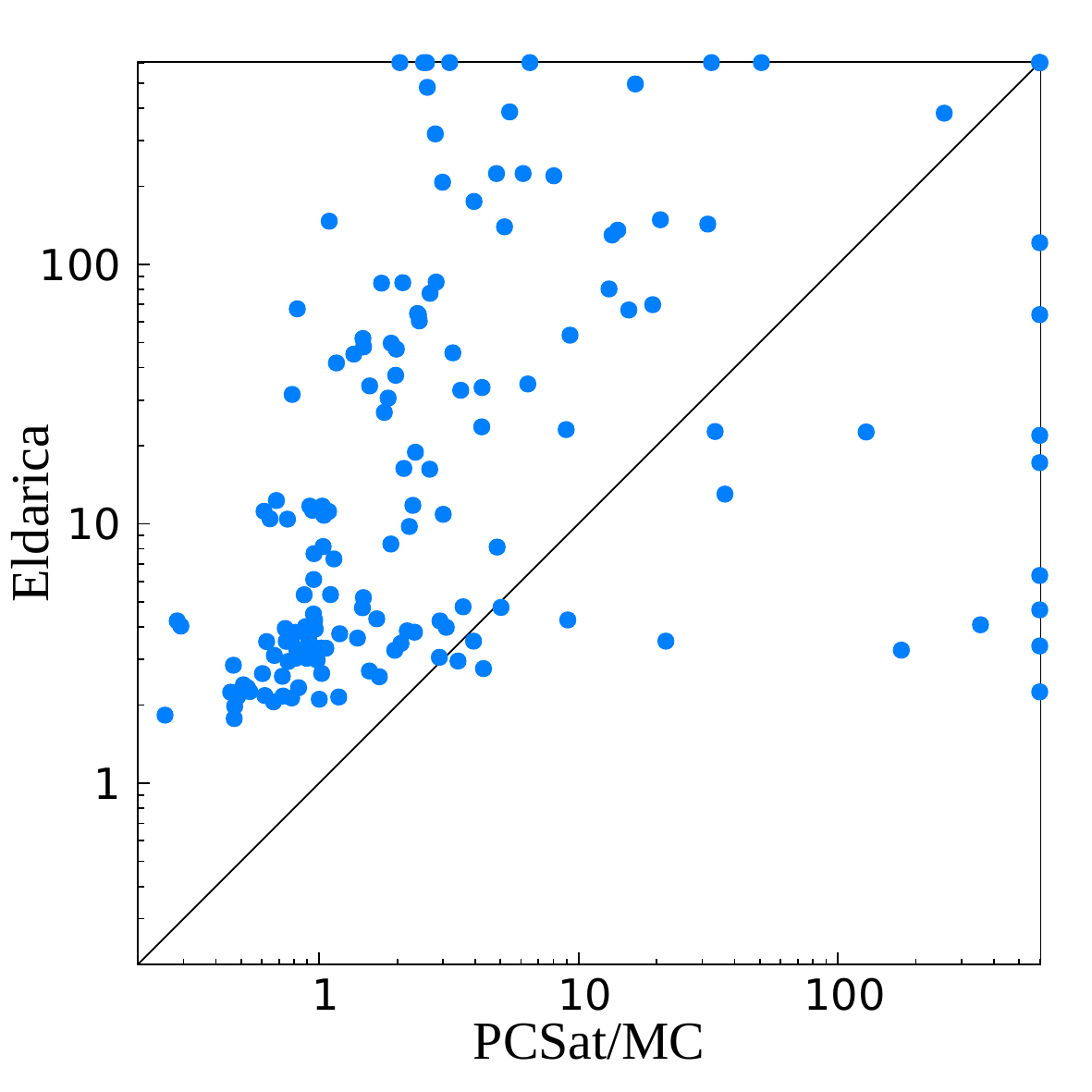}
 }
 \subfloat[\texttt{joint} vs. \texttt{expert}]{
   \includegraphics[width=0.32\textwidth]{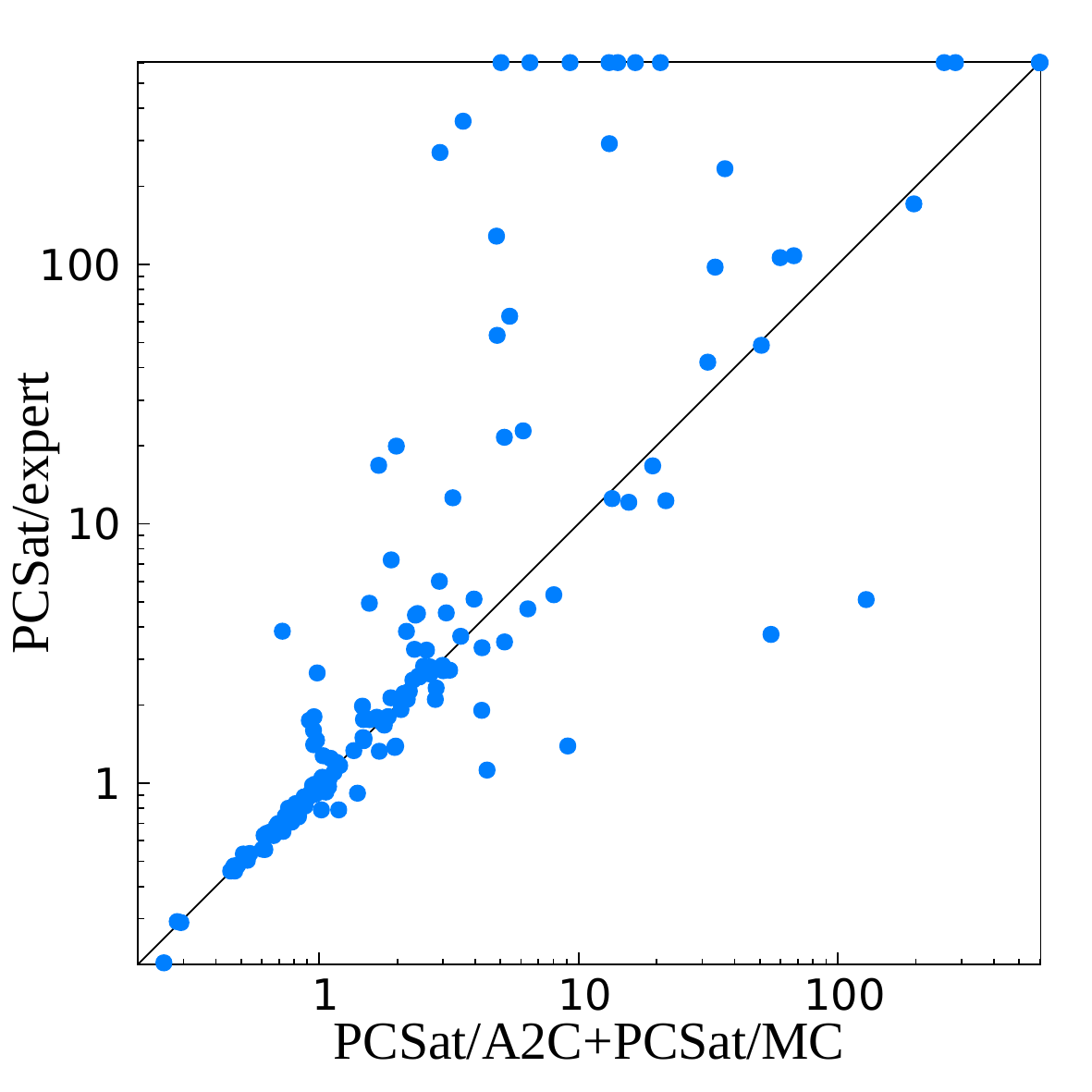}
 }
\caption{Comparison of the time spent on each problem by different methods. The scale is logarithmic.}
 \label{fig:baselines-scatter}
\end{figure*}

The difference in the number of solved problems between the two learned heuristics seems to be minimal. Nonetheless, we observed that \texttt{PCSat/A2C} solves three problems whose solutions were not found by \texttt{PCSat/MC} and \texttt{PCSat/MC} solved four problems not solved by \texttt{PCSat/A2C}. In fact, as illustrated by \hyperref[fig:baselines-scatter]{\autoref{fig:baselines-scatter}a}, while the two policies are similar in terms of time spent on solving most of the problems, they start to disagree on some of the problems when a larger time limit is allowed, with one being able to solve them in a short period of time, and the other taking significantly longer. This suggests that the difference in learning algorithms may have indeed induced different heuristics, and using learned heuristics jointly may boost the performance of PCSat in practice.

\begin{figure*}[t] \centering
   \includegraphics[width=1.0\textwidth]{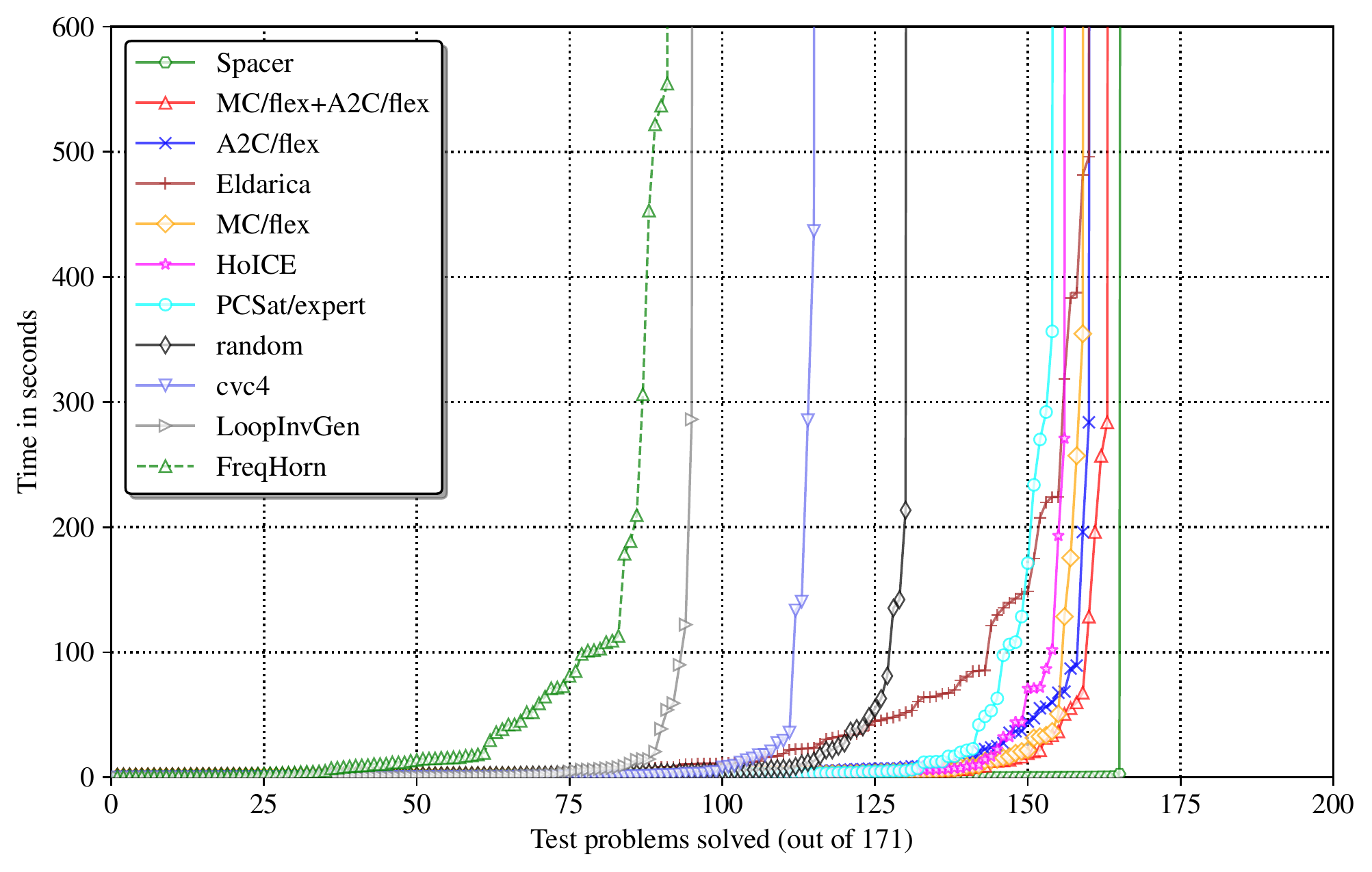}
\caption{Comparison of the cumulative time of each method.}
 \label{fig:baselines-cactus}
\end{figure*}

PCSat guided by the learned heuristics outperforms most of the existing non-learning-based solvers given a time limit of 60 seconds, with the only exception of Spacer. The difference in performance between the solvers is reduced when a larger time limit is allowed. \autoref{fig:baselines-cactus} and \autoref{tab:baselines-600s} show the difference in performance when using a time limit of 600 seconds. It can also be seen that the learned heuristics achieved the best performance among all CEGIS-based solvers regardless of the time limit. Spacer is the only one that significantly outperforms PCSat with learned heuristics and every other solver. The performance of Spacer on the test set seems to be an outlier, for which we do not know the exact reason.
Nonetheless, below we are forced to use a different test set in order to compare the performance of our framework with that of Code2Inv, and noticed that Spacer is not always the winner on every benchmark.
%

\paragraph{Comparison with Code2Inv.}
We now compare the learned heuristics with the neural solver Code2Inv. Ideally, we would like to evaluate Code2Inv on our test set. However, the tools accompanied by Code2Inv for generating the graph representations of given problems failed to correctly convert the problems in our benchmark into a form readable by Code2Inv. For fairness, we use all the common problems that occur in both our data set and the Code2Inv benchmark, in their original forms readable by each solver respectively. There are 81 such problems in total. Since our training set may contain problems from the Code2Inv benchmark, we re-train our agent on a training set containing 127 problems from the SyGuS-Comp 2018 competition, which does not overlap the set of 81 common problems.

\begin{figure*}[t] \centering
 \subfloat[\texttt{A2C} vs. \texttt{Code2Inv}]{
   \includegraphics[width=0.45\textwidth]{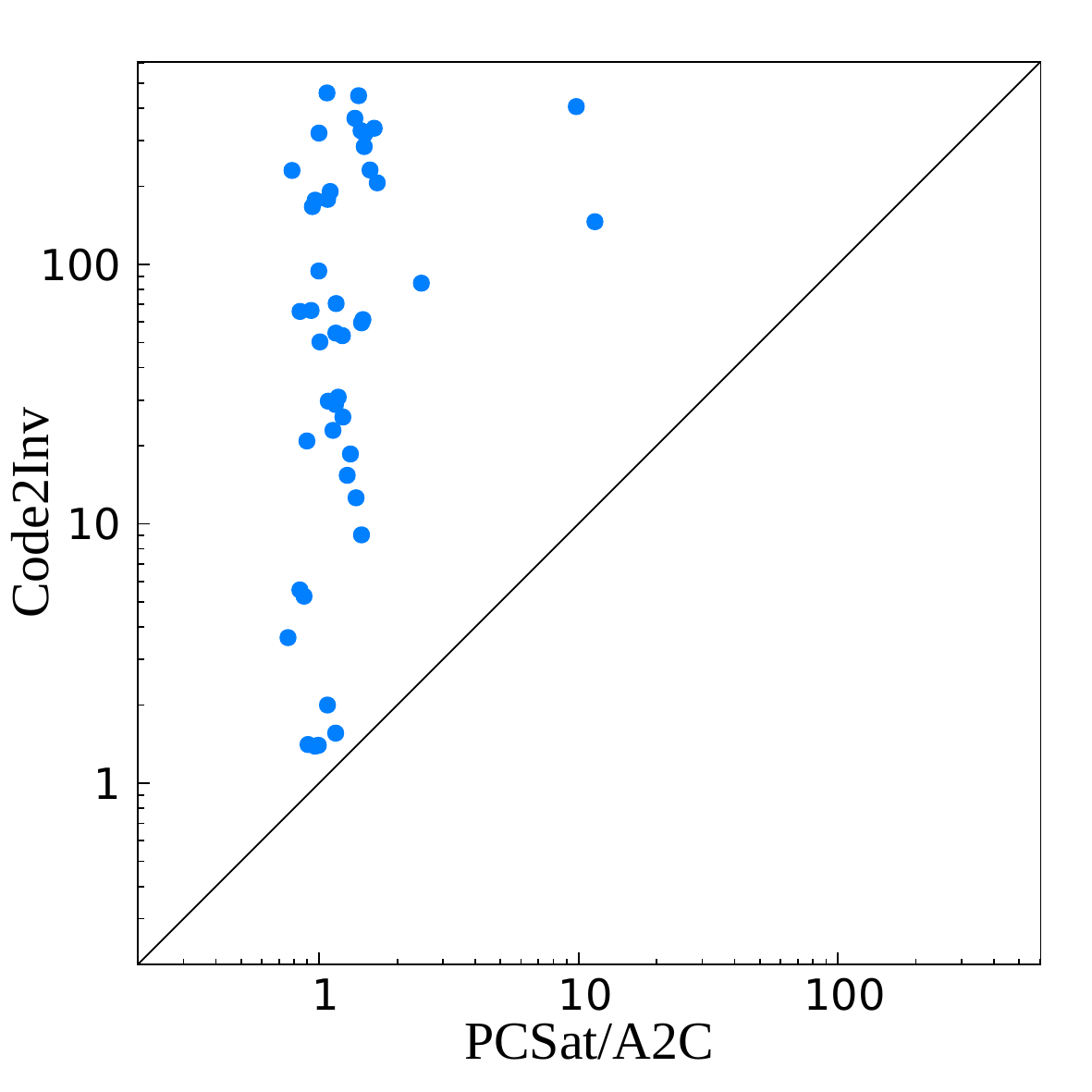}
 }
 \subfloat[Cumulative time]{ \includegraphics[width=0.45\textwidth]{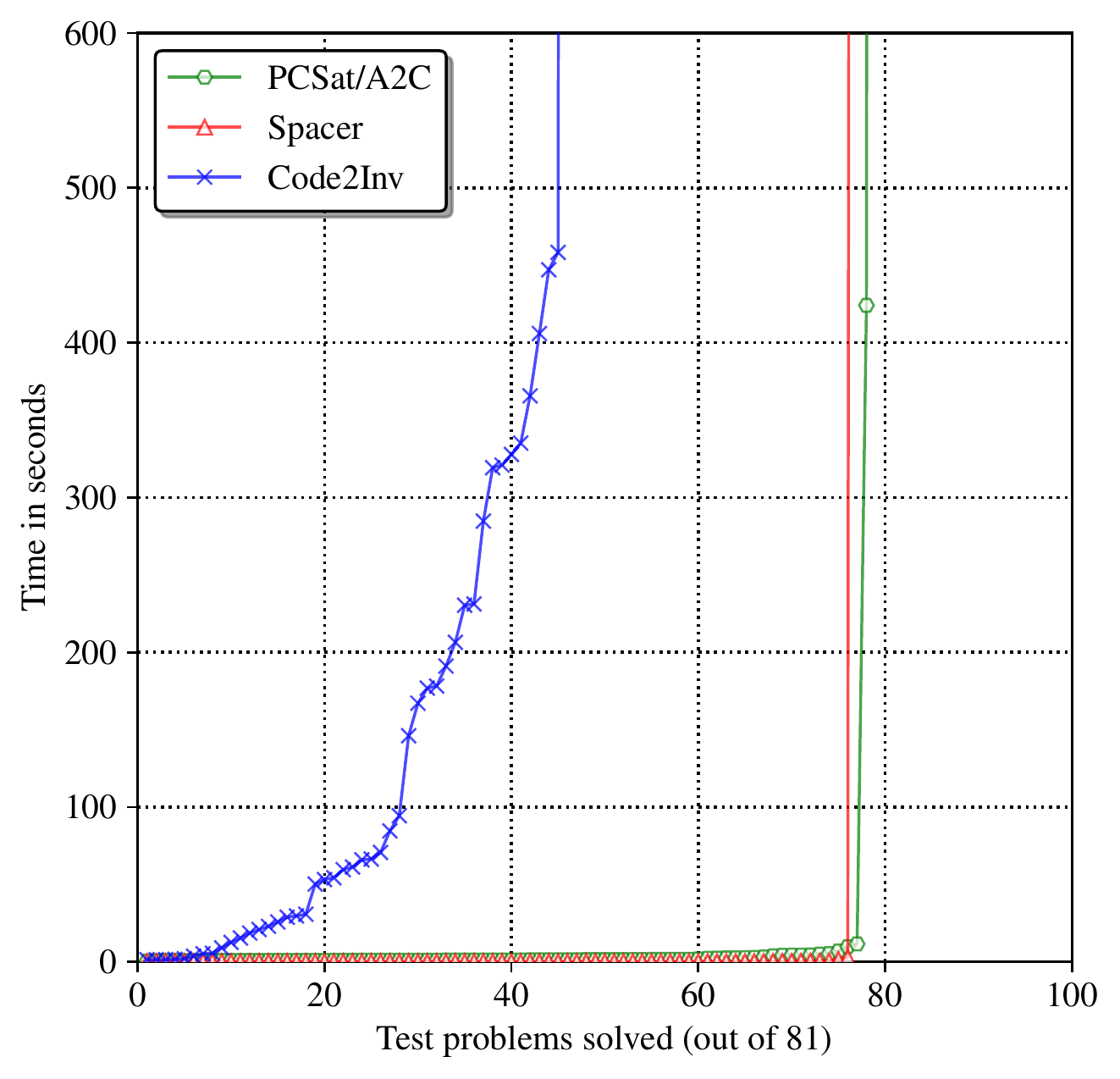}
 }
\caption{(a) Comparison of the time spent on each problem by Code2Inv and PCSat using our learned heuristics. (b) Comparison of the cumulative time of each method.}
 \label{fig:code2inv}
\end{figure*}

\autoref{tab:code2inv} shows the performance of our agent and that of Code2Inv as an out-of-the-box solver given a time limit of 600 seconds. We also add the performance of Spacer as an indicator of the state-of-the-art performance on this set of problems. It can be seen that PCSat with learned heuristics not only outperforms Code2Inv by  solving 50\% more problems, but also slightly surpasses Spacer in terms of the number of solved problems. \hyperref[fig:code2inv]{\autoref{fig:code2inv}a} and \hyperref[fig:code2inv]{\autoref{fig:code2inv}b} illustrate the performance on individual problems. Noticeably, every problem solved by Code2Inv is solved by our agent in a shorter time. Our agent is also able to find four solutions that were not found by Spacer.

The limited performance of Code2Inv is possibly due to the need to learn every problem from scratch, because the shapes of the neural networks in Code2Inv's algorithm depend on the graph representation of the target problem. Different problems may use different neural networks, which makes it difficult to apply the learned policy for one problem to another which has a completely different graph representation. In contrast, our approach does not have such a restriction --- the learned policies are applicable to all the problems as long as the state and action spaces remain unchanged.

\begin{table}[h]
  \caption{Performance of PCSat guided by the learned heuristics and that of Code2Inv and Spacer, on the new test set.
\\~}
  \label{tab:code2inv}
  \centering
\begin{tabular}{lccccc}
\toprule
 Methods & approach & ~sat~ & ~unsat~ & ~timeout~ & ~time(s)~ \\
    \midrule
\texttt{Code2Inv} & CEGIS & 45 & - & 36 & 27289 \\
\texttt{Spacer} & PDR & 70 & 6 & 5 & 3009\\
\midrule
\texttt{PCSat/A2C} & CEGIS & 72 & 6 & 3 & 2373\\
\bottomrule
\end{tabular}
\end{table}

\subsection{Ablation: design of the state space}
So far we have been using 
the state space as described in~\Cref{sec:method}. In this section, we study how a different state space may affect the performance of the learned heuristics. In particular, we are interested in knowing whether embedding expert knowledge into the state space helps improve the performance. To see if the expert knowledge helps, we introduce a new design of state space as follows which reflects the human expert knowledge of invariant synthesis.

\TT{added}
Recall that the strategy of PCSat only uses templates of the form \( ([N]*M, P, Q) \) where \( [N]*M \) is the list of length \( M \) consisting only of \( N \).
\TT{this paragraph was slightly changed}
An \emph{expert state} is an abstraction obtained from the original state \( ([N]*M, P, Q, f_1, f_2, z) \). Concretely, an expert state is a tuple $(b_0,b_1,b_2,b_3,b_4,b_5,f_1,f_2,z')$ of the boolean values, each element
of which summarizes a corresponding parameter of the original state as follows.
\begin{description}
\item[$b_0$ (resp.\ $b_1$)] $\Leftrightarrow$ $M < N$ (resp.\ $M = N$).
\item[$b_2$ (resp.\ $b_3$)] $\Leftrightarrow$ $P = \infty \vee P \ge 2$ (resp.\ $P = \infty \vee P \ge 5$).
\item[$b_4$ (resp.\ $b_5$)] $\Leftrightarrow$ $Q = \infty \vee Q \ge 2$ (resp.\ $Q = \infty \vee Q \ge 5$).
\item[$f_1, f_2$] are the same as in the original state.
\item[$z'$] $\Leftrightarrow$ $z > 0$.
\end{description}

This design of the state space is inspired by the heuristic engineered by the developers of PCSat (i.e., the heuristic used in the baseline \texttt{PCSat/expert} in~\Cref{sec:evaluation}). For example, the boolean value $b_1$ can be used to alternate
incrementing $M$ and $N$. The baseline heuristic also uses the values of $P$ and $Q$
to increase $P$ and $Q$ gradually, and the developers of PCSat believe that 2 and 5 are
good magic numbers that balance the expressiveness of the templates well.
The developers also believe that
whether \( z=0 \) or not should be the most significant information about \( z \).
\TT{this sentence was changed}

The expert state space is much smaller than the original state space as it has eliminated ``irrelevant'' possible states as believed so by human experts. The size of the expert state space is only 512. Such a state space is especially useful when training with tabular methods such as MC, as we do not need to worry about huge state-action value tables that are impossible to be fully explored. Given the expert state space, the corresponding \emph{expert action} is then represented by \( (n,m,p,q) \in \{0,1\}\times\{0,1\}\times\{0,1,\infty\}\times\{0,1,\infty\} \), which updates \( ([N]*M, P, Q) \) to \( ([N+n]*(M+m), P+p, Q+p) \).

\TT{changed: "raw flexible-template encoding" \(\to\) "raw states"}
We train an agent using MC with the expert state space and expert action space, and compare its performance with
that of the agents trained with the original state space defined in \Cref{sec:method}. We call states in the original state space \emph{raw states} below. 
\autoref{tab:ablation} shows the performance of the four agents. It can be seen that while MC using the expert state space learns well, the extra expert knowledge does not help solve more problems. Conversely, the slightly worse performance may be a result of the overly abstracted state space --- an agent using this space may not be able to distinguish states that may have otherwise led to different updates of a template when using a richer representation of states such as the raw states. 

\begin{figure*}[t!] \centering
 \subfloat[\texttt{raw} vs. \texttt{expert}]{

   \includegraphics[width=0.45\textwidth]{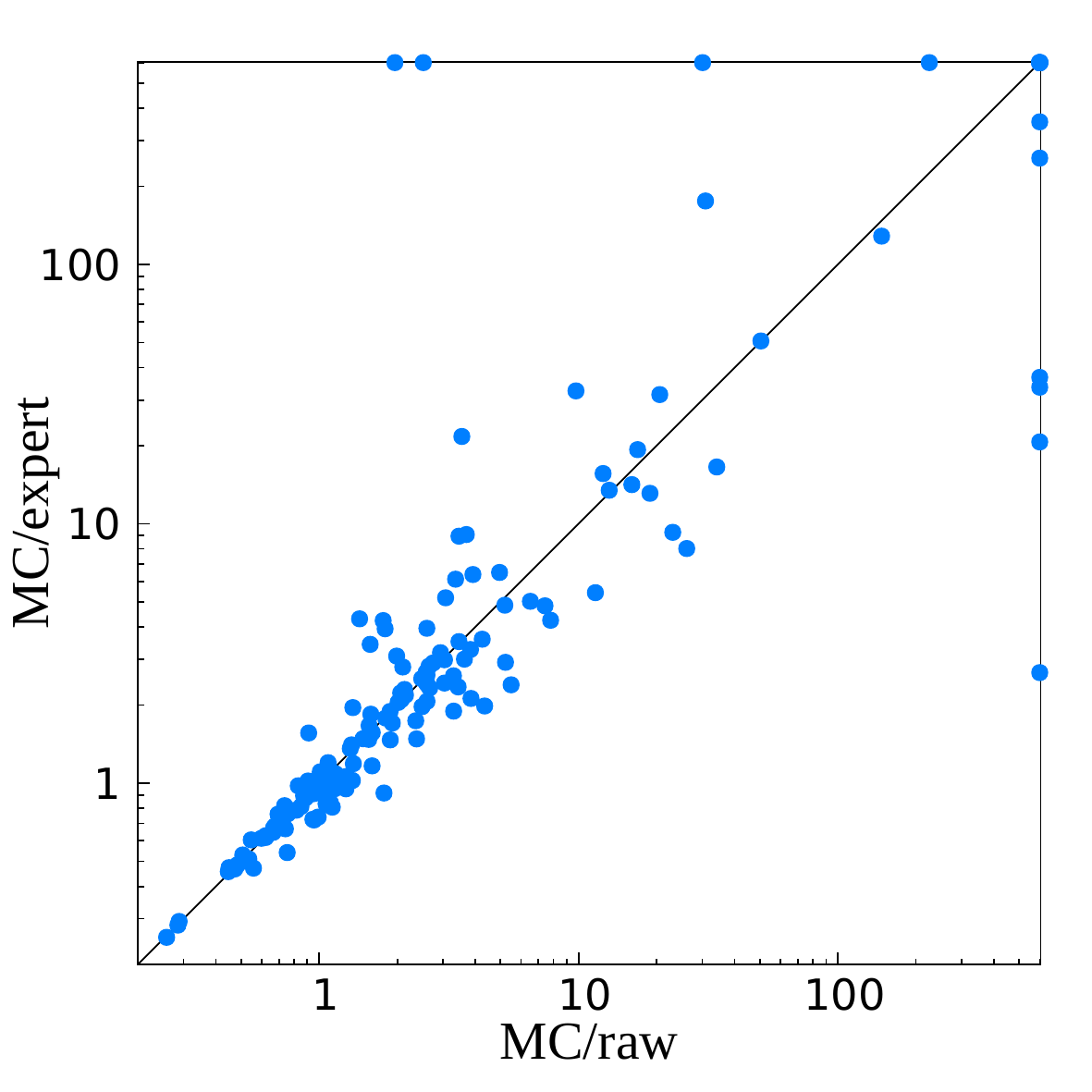}
 }
\subfloat[\texttt{joint} vs. \texttt{joint}]{
   \includegraphics[width=0.45\textwidth]{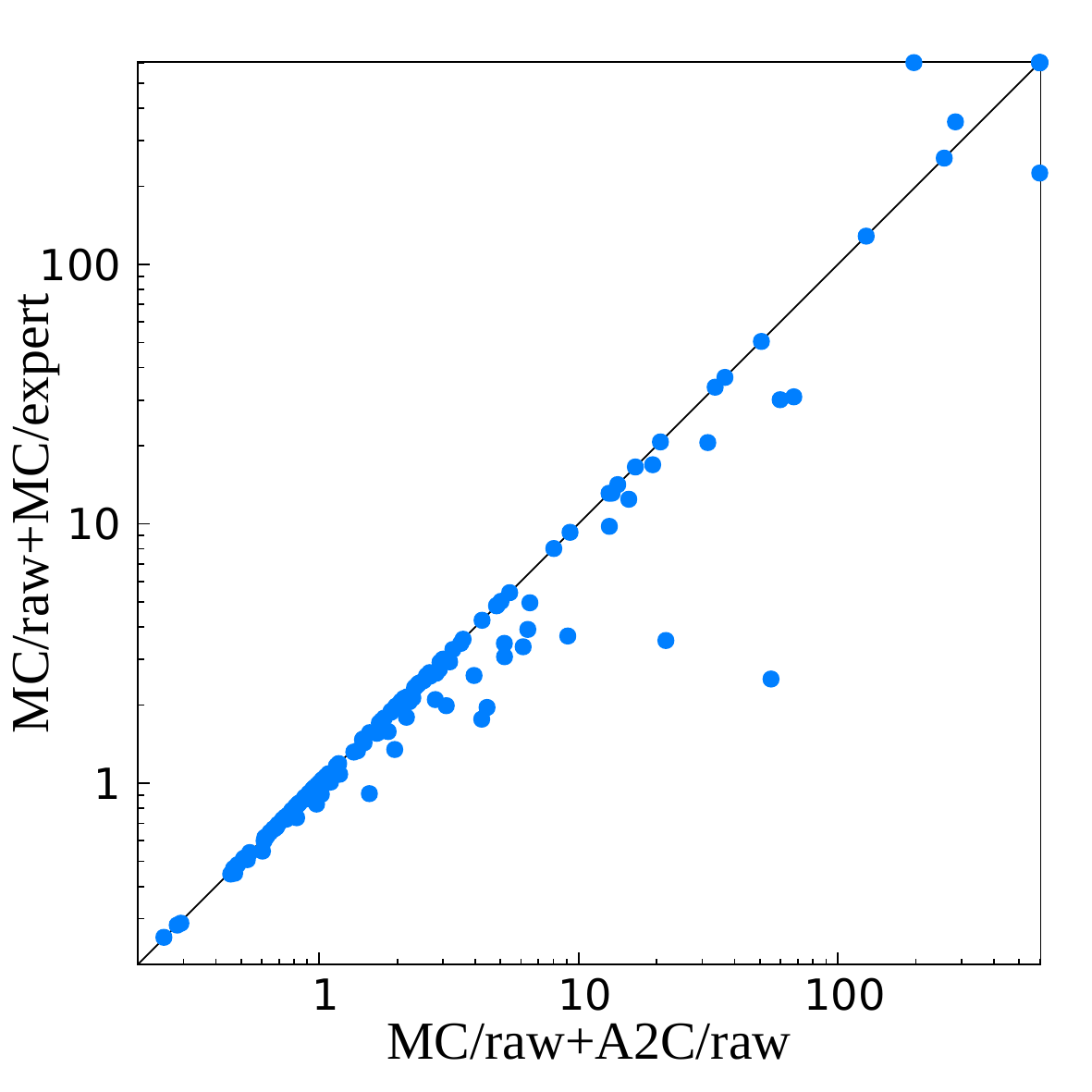}
 }
 \\
\subfloat[Cumulative time]{
   \includegraphics[width=0.7\textwidth]{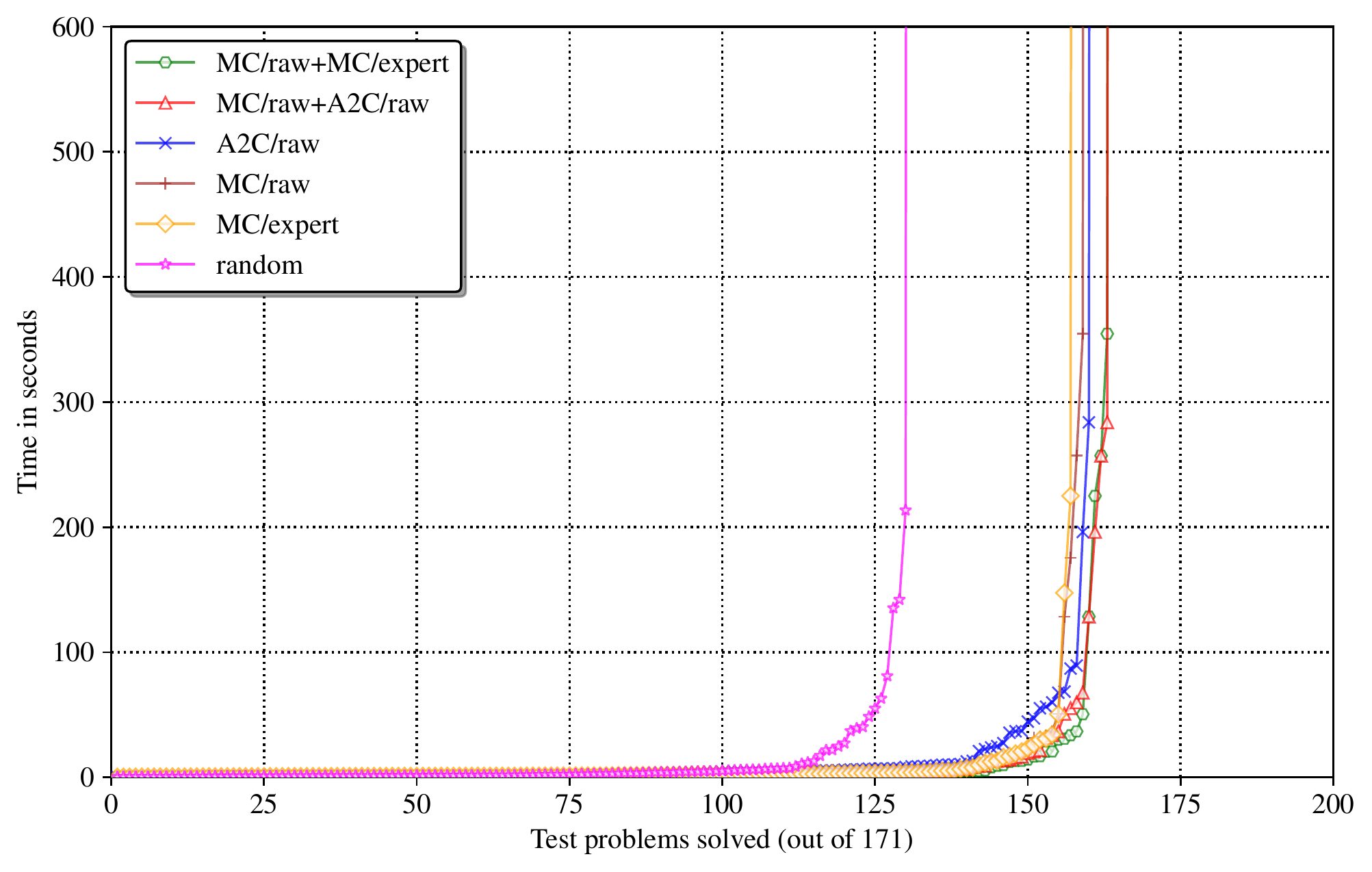}
 }
\caption{(a)(b) Comparison of the time spent on each problem by different methods. (c) Cumulative time of each method.}
 \label{fig:ablation}
\end{figure*}

On the other hand, \hyperref[fig:ablation]{\autoref{fig:ablation}a} shows that MC using expert states helped find solutions not found by MC using the raw states, and vice versa. \hyperref[fig:ablation]{\autoref{fig:ablation}b} and \hyperref[fig:ablation]{\autoref{fig:ablation}c} further shows that the performance of \texttt{MC/raw} working together with \texttt{MC/expert} is almost identical to that of it working together with \texttt{A2C/raw}, suggesting that the expert insight embedded in the expert states might be learnable by A2C using the raw states.

\begin{table}[h]
  \caption{Performance of the learned heuristics using the expert states, as indicated by \texttt{MC/expert}, and the learned heuristics using the raw 
  states, as indicated by \texttt{MC} (resp.\ \texttt{A2C})\texttt{/raw}.
\\~}
  \label{tab:ablation}
  \centering
\begin{tabular}{lcccc}
\toprule
 Methods &   ~sat~ & ~unsat~ & ~timeout~ & ~time(s)~ \\
    \midrule
\texttt{random} & 121 & 9 & 41 & 25858\\
\texttt{MC/expert} & 148 & 9 & 14 & 9337\\
\texttt{MC/raw} & 150 & 9 & 12 & 8705\\
\texttt{A2C/raw} & 151 & 9 & 11 & 8318 \\
\bottomrule
\end{tabular}
\end{table}

%% file: related.tex
%
%
%
%


This paper focused on a template-based approach to CEGIS~\cite{Garg2014,Alur2015,Unno2021}, while grammar-based synthesis~\cite{Fedyukovich2018a,Si2018,Si2020,Kalyan2018,Bunel2018,Chen2020} (e.g., Code2Inv)  and decision-tree learning~\cite{Krishna2015,Garg2016,Champion2018,Zhu2018,Ezudheen2018,Kura2021} (e.g., HoICE) are also popular approaches to guessing a candidate solution from gathered data in CEGIS.
Although the template-based approach is advantageous in that it
adaptively adjusts atomic formulae to be used in a candidate solution, it requires careful tuning of a heuristic for deciding the
shape of a candidate solution.  We have addressed this challenge by leveraging RL to learn an effective heuristic.

Si et al.~\cite{Si2018,Si2020} proposed a framework called Code2Inv to learn
loop invariants with deep reinforcement learning.  Code2Inv uses
graph neural networks to encode the information of a program and
synthesizes loop invariants using a syntax-directed encoder-decoder structure.
In contrast to their approach that tries to
synthesize an invariant directly from a program, we learn the heuristics that guide the search in existing tools for program verification. One restriction of Code2Inv is that the shape of the neural networks that parameterize its policy depend on the graph representation of the target problems. It is unclear how a policy learned for one problem can be easily applied to another that has a completely different graph representation. In contrast, our policy learned for heuristics is universal --- it works with any problem readable by the base solver.
%
%

Code2Inv and our approach also differ in the evaluation of the obtained
solver.  Si et al.\ evaluated the performance of their solver in the
numbers of queries to Z3~\cite{Moura2008},
instead of measuring the actual running time.  Although the running cost of Z3 is
dominant in most of the program verification tasks, Code2Inv conducts
online learning of a neural network, which incurs non-trivial
additional cost to the performance.  On the contrary, the performance
of our solver is evaluated in the wall-clock time.

RL has also been applied to program synthesis~\cite{Bunel2018,Chen2020} and relational verification~\cite{Chen2019}.
These works learn heuristics that select one inference rule or production rule from finitely many options, while we apply RL to aid prioritized search for an infinite set of candidate solutions. In other words, we learn heuristics to synthesize cut-formulas, which correspond to loop invariants in our problem setting. Approaches such as Concord-StandardPG~\cite{Chen2020} also need to retrain the policies for each problem just like Code2Inv~\cite{Si2018,Si2020}, while our approach does not have such limitation.

\HU{We better compare with them in more details.}


%
The benefits from learning heuristics for theorem proving has been
demonstrated in various automated theorem proving techniques including
CDCL for SAT~\cite{Liang2018,Selsam2019,Selsam2019a}, strategies for
SMT~\cite{Balunovic2018}, connection tableau~\cite{Kaliszyk2018}, and
incremental determinization for QBF~\cite{Lederman2020,Selsam2019a}.
These previous studies applied machine learning to enhance the proof
search.
We cannot directly apply these theorem proving techniques to the ISP because we
need to synthesize an appropriate predicate as a solution to the ISP, in
addition to deciding the validity of a given formula.
%
%
%
%

%
Extensive research has been conducted on learning embedding of
programs and logical formulae through neural networks such as
LSTMs~\cite{Iyer2016,Irving2016,Kaliszyk2017}, tree-based neural
networks~\cite{Mou2016,Loos2017,Evans2018,Sekiyama2018}, graph neural
networks~\cite{Allamanis2018,Wang2017,Kusumoto2018,Paliwal2020}, and a
path-based attention model~\cite{Alon2019}.
The present work does not use learned embedding of invariants or
programs.  It is an interesting future direction to investigate
whether using embedding serves for any further improvement.




%% file: conclusion.tex
\section{Conclusion}
\label{sec:conclusion}

We presented how to apply reinforcement learning to the task of learning effective heuristics for a template-based CEGIS procedure.
%
To this end, we modeled the behavior of the procedure as an MDP and a heuristic as an agent that updates a template.
We trained the agents using the first-visit on-policy Monte Carlo control (MC) and Advantage Actor-Critic (A2C); the learned heuristics are the best in its kind --- they outperformed the heuristics engineered by human experts, and achieved the best performance among the CEGIS-based solvers. The learned heuristics have also demonstrated comparable (and sometimes superior) performance to that of the state-of-the-art non-CEGIS-based solvers, validating the effectiveness of our approach.
%
%


We have focused on programs that have only one loop, whose loop-invariant synthesis can be reduced to solving a linear CHC with a single predicate variable.
Future work includes handling a broader class of constraints such as CHCs with multiple predicate variables for safety verification of a program with multiple loops and the class pfwCSP of constraints for relational~\cite{Unno2021} and branching-time temporal~\cite{Unno2020} verification.




